\documentclass[11pt,twocolumn]{article}

\usepackage[utf8]{inputenc}
\usepackage{cmap}
\usepackage{graphicx}
\usepackage{amsmath}
\usepackage{amssymb}
\usepackage{hyperref}
\usepackage{xspace}
\usepackage{float}
\usepackage[left=2cm,right=2cm,top=1.5cm,bottom=2cm,bindingoffset=0cm]{geometry}
\usepackage{breakurl}

\hypersetup{
  breaklinks=true,
  hidelinks = true,
  pdftitle = {A Multi-head-based architecture for effective morphological tagging in Russian with open dictionary},
  pdfauthor = {Skibin K., Suschenko S., Pozhidaev M.},
  pdfkeywords = {Morphological tagging, Multi-head attention, Transformer, Embedding, Positional encoding},
  pdfcreator = {},
  pdfproducer = {}
}

\newcommand {\titlefont} {\Large \bf \sffamily}
\newcommand {\authorfont} {\large \sffamily}
\newcommand {\publisherfont} {\normalsize \sffamily}
\newcommand {\abstractfont} {\normalsize \it}

\begin{document}
\sloppy
\large

\begin{center}
  {\titlefont
    A~Multi-head-based architecture for~effective morphological tagging in~Russian with~open dictionary
  }

  {\authorfont
    Skibin~K.,
    Suschenko~S., % Добавлен пробел перед инициалом 
    \href{https://marigostra.com}{Pozhidaev~M}.
  }

  \vspace{0.5cm}

  {\publisherfont
    \href{https://en.tsu.ru/}{National Research Tomsk State University}\\
    \href{https://ainlconf.ru}{Submitted to~AINL-2026}\\
      }

  \end{center}

  \vspace{0.5cm}

  {\abstractfont
  The~article proposes a~new~architecture based on~Multi-head attention to~solve the~problem of~morphological tagging for~the~Russian language.
  The~preprocessing of~the~word vectors includes splitting the~words into subtokens, followed by~a~trained procedure for~aggregating the~vectors of~the subtokens into vectors for~tokens.
  This allows to~support an~open dictionary and analyze morphological features taking into account parts of~words (prefixes, endings, etc.).
  The~open dictionary allows in~future to~analyze words that are absent in~the~training dataset.

The~performed computational experiment on~the~Sintagrus and Taiga datasets shows that for~some grammatical categories
the~proposed architecture gives accuracy 98--99\% and above, which outperforms previously known results.
For~nine out of~ten words, the~architecture precisely predicts all grammatical categories and indicates when the~categories must not be analyzed for the word.

At~the same time, the~model based on~the proposed architecture can be trained on~consumer-level graphics accelerators,
retains all the~advantages of~Multi-head attention over RNNs (RNNs are not used in~the~proposed approach),
does not require pre-training on large collections of unlabeled texts (like BERT),
and shows higher processing speed than previous results.
  }

  \section*{Introduction}

Morphological tagging is the process of automatically identifying and classifying grammatical characteristics of words in a text.
It remains one of the key tasks in the field of computational linguistics and its application in software engineering.
The successful solution of this problem can have great importance both in fundamental linguistics,
because in many cases it turns out to be inextricably related to~the~solution of the problem of homonymy removal,
and to a number of purely practical applications, including machine translation, sentiment analysis, information extraction, identification of linguistic patterns, etc.,
because it allows to~clarify the meaning and content of the text.
The problem of homonymy is particularly acute in creating audiobooks,
since the ambiguity of grammatical attributes leads to frequent errors in stress positioning.

Solving this problem by a human often involves understanding the meaning of a text,
because it often requires determining the grammatical characteristics of fictional words that are absent in dictionaries.
Such a task arises when processing fiction texts in the fantasy genre,
which are characterized by the authors' desire to build fictional worlds.
The possibility of processing fictional words will be further denoted by the term \textit{open dictionary}.

The development of AI technologies can make a new significant  contribution to solving this problem.
In this work, special attention is paid to the idea of using the multi-head attention (MHA) architecture (Vaswani, 2017 \cite{transformer}) in its plain form,
i.e. without building a transformer for~training on~large texts collections,
which can be observed in previous solutions based on the BERT model (Devlin, 2019 \cite{bert}).
The use of a plain variant of~the~MHA architecture makes it possible to significantly reduce the need for computing resources
that can be faced when using BERT,
make it possible to train a~custom model on user data using consumer-level graphics accelerators.
This also provides support for an open dictionary without noticeable losses in the quality of solutions compared to the best results
obtained for the Russian language.

\section{Related Work}

The problem of morphological tagging in computational linguistics has a long history of research,
which can be divided into three stages.
The following three stages are the~approaches based on statistical methods or neural networks.
The~approaches based on analytical grammars or any other formal structures, including dictionaries, are not considered there due to their extremely limited capabilities and narrow scope of application.

At the first stage, general statistical methods were used, including hidden Markov models (HMM, Rabiner, 1986 \cite{Rabiner} ). % Убран пробел перед точкой
In some of them, the tokens of the source text were declared observable states,
and the corresponding grammatical attributes were hidden states.
After that, it's  necessary to use the Viterbi algorithm (Viterbi, 1967 \cite{Viterbi}) to determine the most likely sequence of hidden states based on statistical data obtained from the analysis of marked-up text bodies for the Russian language.
The disadvantages of this method include its inability to work with an open dictionary.

The second stage is characterized by the use of neural networks before the appearing of the transformer architecture.
At this stage, attempts were made to use recurrent neural networks (RNN, Elman, 1990 \cite{Elman}).
RNN's are usually used in~conjunction with memory models,
including LSTM (Anastasiev et al, 2018 \cite{lstm})
or GRU (Movsesian, 2022 \cite{gru}).
RNN's have significant drawbacks comparing with transformer-based architectures,
which authors made major effort to~eliminate them from NLP tasks.
Two disadvantages of~RNN's should be especially highlighted:

\begin{itemize}
\item{Sequential operation mode: each subsequent element in the ~RNN is processed only after the previous one,
what makes parallel data processing difficult and prevents the effective use of modern graphics accelerators}
\item{Unequal influence of the elements at the ~edges of the input sequence compared to the ~elements in the~middle part of it.}
\end{itemize}

The third stage, which is the most sophisticated, includes various attempts to apply various forms of the transformer architecture.
In~particular, the BERT model.
This is a class of the most advanced algorithms that show the state of the art level of quality, but at the same time, it often remains extremely resource-consumptive both in the sense of computing resources and in the sense of available training data.

In this category of models, the morphological tagging tool based on the BERT model, proposed by the MIPT DeepPavlov laboratory \cite{dp}, should be noted.
Our research shows that the use of the BERT model itself  has a~
moderate effect on the result comparing to the plain MHA,
while BERT assumes a long training procedure on an unlabeled corpus of texts using industrial graphics accelerators, not needed with the plain MHA.

\section{Model Description}

\subsection{Tokenization}

Before discussing the architecture of the model and the~calculation pipeline,
we should choose the~tokenizer.
As in any other linguistic algorithm or deep learning linguistic model,
the tokenizer plays an important role in the described here architecture,
splitting text data into smaller elements.

An obvious solution at first glance, in which the ``one token - one word'' rule is applied,
cannot
be applied because the architecture will not be able to process words
that were missing at the stage of tokenizer training.
In this case, we would get an architecture with a closed dictionary,
which we consider to be a~significant drawback,
which can be observed, for~example, in~the Word2vec architecture (Mikolov, 2013 \cite{word2vec}).

Having accepted the inadmissibility of the formation of tokens on ~word boundaries,
two approaches were considered instead:
Byte-pair Encoding (BPE) (Kozma, 2024 \cite{bpe}) and $n$-grams like in the Fasttext architecture (Bojanowski, 2016 \cite{fasttext}).
Both of these options allow for text analysis with ~open dictionary support.

BPE is a data compression algorithm that is based on the idea of sequentially replacing the most common byte pairs with a new, previously unused byte. 
BPE starts by analyzing the frequency of occurrence of all possible byte pairs in the original dataset and then iteratively replaces the most frequently occurring pairs with a new character. 
The process continues until a specified stopping criterion is reached, for example, a certain dictionary size or data compression ratio.

It's necessary to keep in~mind  that the~quality of~morphological analysis may potentially depend on how the word is broken down into tokens.
The~BPE tokenization algorithm was chosen because of~its flexibility.
It offers the possibility of choosing the threshold for the occurrence of neighboring tokens to merge them into one token.
This parameter can be used to adjust both the size of the dictionary and the size of tokens (in terms of the number of characters).
The lower the value of the token occurrence threshold,
the~more tokens a~word will be split into.
If the parameter value is high, the tokens tend to~be whole words.

To~study the~risk of~influence of~BPE on~quality,
seven morphological classifier models with different parameters of the threshold frequency of token merging were trained.
Since the minimum frequency of tokens appearing together to combine them is a relative value and depends on the training data,
we will provide values for the dataset used for training and testing the model.
Seven values of the threshold frequency of merging were selected for testing.: 500, 600, 800, 1000, 1200, 1400, 2000.
As~a result, they all showed approximately the same results.
The~discrepancy of accuracy in predicting all grammatical attributes of a word and accuracy in predicting the entire sentence is no more than 1\%.
There was no strict linear relationship between the quality and the value of the minimum frequency of fusion.

As a~part of a computational experiment, we tried to determine the optimal value of the dictionary length,
checking its influence on~the quality of the model.
The~results showed that for ~Russian,
the~best behavior of the model is achieved with ~a dictionary length of 2198 tokens.

The $n$-gram tokenizer from the FastText model is an algorithm for splitting text into tokens
using a pre-trained vector representation model of words. 
$n$-grams are all possible substrings of a certain length $n$ extracted from a word. 
This approach is considered as an additional option for the development of our architecture  for~further research.

\begin{figure*}[tb]
    \centering
\includegraphics[width=0.9\textwidth,keepaspectratio]{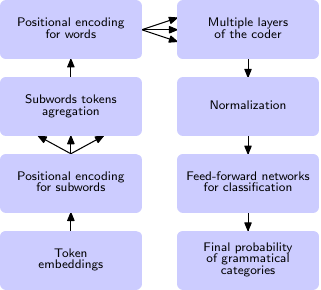}
    \caption{Data processing pipeline}
    \label{arch_fig}\vspace{-.3cm}
\end{figure*}

\subsection{Word representation}

As mentioned in the previous section,
a word is represented not by a single token, but by a sequence of tokens.
Without~this feature, open dictionary support would not be possible.

The representation of a word by several tokens led to~two features in~the implementation of the proposed architecture:

\begin{enumerate}
\item{
It is required to limit the length of each word in ~tokens to a certain fixed value.
A length value of six corresponds to a quantile of 0.98 in ~a fraction of words.
It was selected, and all words with a length of more than six tokens were cut off.
}
\item{
  When calculating gradients on the reverse pass of~the model operation with the error backpropagation,
  the vectors included in the corresponding word were summed.
}
\end{enumerate}

Since we are solving the task of classifying the attributes of words, when the model is trained, each word is assigned the correct value of the grammatical attribute,
which is compared with the one obtained by the model during the forward pass.
The multiplicity of tokens for ~word representation requires us, in addition to the approach proposed above, to consider the~alternative approaches:

\begin{enumerate}
\item{
Assign a grammatical attribute to only one word token. For example, only the last one or only the first one.
}
\item{
Assign a grammatical attribute to all tokens of a word.
}
\end{enumerate}

These two approaches were not considered, since the first would obviously lead to loss of information, and the second would lead to its unjustified mixing.

\subsection{Basic Encoding of segments}

At ~each step, the architecture performs processing  of the four-dimensional tensor:

\begin{itemize}
\item{
a~set of segments for~one batch
}
\item{
words in each segment
}
\item{
tokens in ~each word
}
\item{
vector representation of each token in the~form of~an~embedding
}
\end{itemize}

The general calculation procedure can be described as follows:

\begin{enumerate}
\item{
The positional encoding (RoPE) is applied to each token inside each word.
}
\item{
The dot product attention is calculated inside each word for ~tokens included in~this word.
Thus, if there are $n$ words in the ~segment, then dot product attention is calculated $n$ times.
}
\item{
Calculation of the score of the token (contribution of the token to the overall representation of the word). To do this, the results of the previous step are processed using a fully connected network with subsequent processing by the softmax() function.
}
\item{
The vectors of token representation within a single word are aggregated into a common vector of word representation by summation with multiplication by the coefficient obtained at the previous step.
}
\item{
The application of positional encoding is repeated once again, but this time at the word level.
}
\item{
The resulting tensor is processed by four transformer encoder blocks in their original form.
}
\item{
The obtained encoding results are processed by a feed-forward network with ~ one hidden layer, which plays the role of a classifier, assigning to~each word an appropriate grammatical feature.
}
\end{enumerate}

\section{Dataset}

  Two datasets were used for training and validation:
  UD SynTagRus ver.~2.16 \cite{Sintagrus} and
  UD Taiga ver.~2.17 \cite{Taiga}.

  Here are the characteristics for each dataset:

  Information about   SynTagRus samples:
  
  \begin{itemize}
 \item { 
Train:
69629 sentences  total,
1205821 words total,
average sentence length is  17.32 words
}
\item {
Dev:
 8906 sentences total,
153585 words total,
average sentence length is  17.25 words
}
\item {
Test:
 8800 sentences total,
157978 words total,
average sentence length is  17.95 words
}
\end{itemize}

  Information about   Taiga samples:
  
\begin{itemize}
  \item {
Train:
119407 sentences total,
1717332 words total,
average sentence length is  14.38 words
}
\item {
Dev:
1260 sentences total,
15624 words total,
average sentence length is  12.4 words
}
\item {
Test:
1217 sentences total,
15440 words total,
average sentence length is 12.69 words
}
\end{itemize}

The~model has been trained on~the~merged dataset SynTagRus + Taiga:

\begin{itemize}
\item {
Train:
189103 sentences total,
2930617 words total,
average sentence length equals to 15.5 words
}
\item{
Dev:
10166 sentences total,
169209 words total,
average sentence length is 16.64 words
}
\item{
Test:
10017 sentences total,
173418 words total,
average sentence length is 17.31 words
}
  \end{itemize}

  \section{Experiments}

  \subsection{Metrics}

The morphological tagging problem is a~classification problem,
and therefore standard classification metrics are used to~assess the~quality of~its solution.
These include:

\begin{itemize}
\item {{\bf accuracy:} reflects the proportion of correctly classified items among all classified items}
\item {{\bf recall:} represents the proportion of correctly identified positive cases among all actually positive cases}
\item {{\bf precision,} which determines the proportion of correctly classified positive cases among all cases classified as positive}
\item{{\bf F1:} the~harmonic mean between precision and recall,
  allows a~balanced assessment of classifier effectiveness}
  \end{itemize}

\subsection{Experiment Setup}

The proposed architecture was implemented as a~PyTorch-based  library developed in~Python language. 
To~reproduce the results,
the library source code is open and publicly available at~\href{https://github.com/kdskibin/morph\_tagger.git}{github.com{\slash}kdskibin{\slash}morph\_tagger.git}. % Изменен адрес репозиторя (поменялся username)
Training and testing was carried out on nVidia RTX~4090 graphics accelerator.
The~AdamW optimizer is used with $\lambda = 1e-5$.
The~batch size is equal to 96.
The training process includes 35 epochs with a change in the learning rate in three stages (variable value of learning rate we take as a~Precaution measure to~prevent overtraining):

\begin{itemize}
\item{for the first 15 epochs, the learning rate was 1e-4}
\item{followed by 10 epochs of learning rate 5e-5}
\item{then 10 more epochs of learning rate 1e-5}.
\end{itemize}

The~training process takes approximately 8--12 hours depending on~the~accelerator model.
The~resulting model has the~size of 195MB and has  48792985 weight parameters.

\section{Results}

\begin{table*}[tb]
  \centering
  \begin{tabular}{r|ccccc}
    Feature & Accuracy & \begin{tabular}{@{}c@{}} Sentence  \\ accuracy \end{tabular} & Precision & Recall & F1 \\
    \hline \\
 upos & 0.98463 & 0.77898 & 0.96833 & 0.96679 & 0.96684  \\
head & 0.91923 & 0.48346 & 0.88273 & 0.88545 & 0.87740  \\
deprel & 0.94271 & 0.47067 & 0.82724 & 0.84452 & 0.82797  \\
Mood & 0.99874 & 0.97944 & 0.94718 & 0.96784 & 0.95057  \\
NumType & 0.99758 & 0.96008 & 0.85935 & 0.90408 & 0.87672  \\
VerbForm & 0.99781 & 0.96325 & 0.98734 & 0.98938 & 0.98805  \\
ExtPos & 0.99828 & 0.97152 & 0.74604 & 0.79496 & 0.75973  \\
Reflex & 0.99974 & 0.99558 & 0.97378 & 0.99896 & 0.98476  \\
Polarity & 0.99443 & 0.91508 & 0.70643 & 0.80416 & 0.73642  \\
Typo & 0.99997 & 0.99954 & 0.97825 & 0.97825 & 0.97825  \\
NameType & 0.99388 & 0.90930 & 0.74352 & 0.73691 & 0.72995  \\
InflClass & 0.99851 & 0.97780 & 0.93738 & 0.84559 & 0.87151  \\
Person & 0.99861 & 0.97610 & 0.99460 & 0.98794 & 0.99062  \\
Poss & 0.99927 & 0.98720 & 0.96347 & 0.99191 & 0.97625  \\
Animacy & 0.98732 & 0.81833 & 0.97048 & 0.96883 & 0.96951  \\
Degree & 0.99115 & 0.86345 & 0.90006 & 0.92511 & 0.90894  \\
Foreign & 0.99947 & 0.99241 & 0.94309 & 0.88598 & 0.90439  \\
Variant & 0.99930 & 0.98816 & 0.98418 & 0.99008 & 0.98686  \\
Number & 0.99097 & 0.86803 & 0.98990 & 0.98949 & 0.98968  \\
Gender & 0.98663 & 0.81431 & 0.98090 & 0.97924 & 0.98000  \\
NumForm & 0.99829 & 0.97180 & 0.94471 & 0.97535 & 0.95617  \\
Aspect & 0.99686 & 0.94701 & 0.98643 & 0.98754 & 0.98691  \\
Case & 0.98395 & 0.79896 & 0.96960 & 0.96860 & 0.96895  \\
PronType & 0.99495 & 0.91836 & 0.90288 & 0.90052 & 0.89366  \\
Tense & 0.99724 & 0.95323 & 0.97064 & 0.97289 & 0.97033  \\
Abbr & 0.99868 & 0.98001 & 0.84428 & 0.89351 & 0.85648  \\
Voice & 0.99579 & 0.92929 & 0.96404 & 0.96103 & 0.96189  \\
      \end{tabular}
      \caption{Quality of~recognizing all grammatical categories.}
  \label{tab:categories}
\end{table*}

The results of the computational experiment of~predicting the~individual categories are shown in~the~Table~\ref{tab:categories}.
The~average accuracy is 99.05233\%.
Since we are unable to find the~same estimation for~BERT-based architecture by~DeepPavlov, we have tried to~make our own run of that model and got the~accuracy 98.98\%.
RNNMorph (Anastasiev, 2018 \cite{lstm}) gives the~accuracy of~prediction  of~upos equal to~98.33\%,
while the~proposed architecture gives better accuracy for~upos equal to~98.463\%.

For~the~case of~predicting all~categories of~a~word the~proposed architecture shows accuracy 95.3559\%.
The~following categories were taken into account: upos, Mood, VerbForm, Person, Animacy, Degree, Variant, Number, Gender, NumForm, Case, Tense, Voice.
For~the~same set of~categories the~GRU-based architecture (Movsesyan, 2022 \cite{gru}) provides accuracy 93.88\%.
The~BERT-based approach shows the~accuracy of~predicting all~categories equal to~97.6\%.
Though  it~is clearly better than obtained by~proposed architecture, we are unable to~be sure that this results can be comparable,
because the~authors do~not give the~exact list of~the~categories being analyzed.
Using different sets of~categories gives different results:
for~example, the~set of categories we mentioned previously (upos, Mood, VerbForm, Person, Animacy, Degree, Variant, Number, Gender, NumForm, Case, Tense, Voice) isn't their full list which can be~found in~the~dataset.
If we test the~full list of~categories, the~accuracy plummets  to~84.861\%.
So this number is~highly dependent on~the~exact list of~categories being tested.

We must admit that in~some situations the~BERT-based architecture by~DeepPavlov can outperforme the~proposed one
and, in~common sense, it is impossible to~clearly state that one of~these architectures is absolutely better than another (though the~difference, very likely, will be  not significant),
but the~proposed architecture requires only 48M of~weight parameters what is clealy less than their number in~BERT  which is  usually equal to~400M and higher.
In~addition, the~whole collection of~training data can be processed on~one consumer-level graphical accelerator and requires approximately 8--12 hours,
what opens new research potential for~institutions without professional-level hardware.
Using the~proposed architecture makes no~dependence on~training collection used for~BERT training.

The~prediction of~all~categories for~all~words in~a~sentence is also highly dependent  of~the~list of~categories being process.
If we take into account absolutely all categories in the~dataset, the~result shows that approximately every fifth sentence (22\%) can be identified fully correctly,
but we still unaware could this result be compared with other results due to unclear list of~categories.

\section*{Conclusion}

An~architecture for~~morphological tagging for~the Russian language based on~the~Multi-head attention,
has been proposed and tested.
It has the~following characteristics: 

\begin{itemize}
\item{
Supports an~open dictionary and is able to~process words absent in~the~~training dataset
}
\item{
Does not~use recurrent calculations and can be used in~highly parallel manner
}
\item{
  For~most grammatical categories, recognition accuracy is 98--99\% or higher,
  and for~the full set of all word categories, the accuracy is close to~90\%
}
\item{
Able to indicate the absence of grammatical categories for~those words for~which they are irrelevant
}
\item{
Performs subtokenization of words and analyzes individual parts of~a~word, calculating their contribution to the encoding vector
}
\item{
Does not~require pre-training on~large unlabeled collections of~text with the use of clusters of industrial graphics accelerators;
}
\item{
Capable of learning and operating on~consumer-level graphics accelerators
}
\item{
Shows an increase in processing speed compared to ~other architectures
}
\end{itemize}

The proposed architecture requires continuing research for~a more precise evaluation of~the control parameters
necessary to~build components based on~it for~practical use.
Areas of further research include testing the architecture for solving the problem of morphological tagging in situations
where the sequence being analyzed is not a~fully written word, but an abbreviated entry
that requires more attention on~context rather than on~internal structure analysis.

\onecolumn % Представление списка литературы в одном столбце

\end{document}